\documentclass[runningheads]{llncs}
\usepackage{graphicx}
\usepackage{orcidlink}
\usepackage{cite}
\usepackage{booktabs}
\usepackage[misc]{ifsym}
\usepackage{amsmath,amssymb}
\usepackage{array}
\usepackage{multirow}
\usepackage{tabularx}
\usepackage{float}
\usepackage{placeins}


\begin{document}
\title{A Dynamic Fusion Large Language Model for Traffic Flow Prediction}
\titlerunning{A Dynamic Fusion LLM for Traffic Flow Prediction}
%
\author{
Xue Qiu\inst{1}\orcidlink{0009-0006-6143-7558}\and
Jianli Xiao\inst{1}\textsuperscript{(\Letter)}\orcidlink{0000-0002-7363-0623}
}
%
%
\institute{
School of Optical-Electrical and Computer Engineering, University of Shanghai for Science and Technology, Shanghai, China \\
\email{233370847@st.usst.edu.cn}, \email{audyxiao@sjtu.edu.cn}
}
\maketitle              
\begin{abstract}
Traffic flow prediction is a core supporting technology for intelligent transportation systems. It uses historical data to infer future traffic dynamics in specific areas, thereby helping to alleviate congestion and improve resource allocation efficiency. Traditional neural networks struggle to break through accuracy limits due to their reliance on singular feature modeling, while large language models (LLMs) suffer from insufficient capture of spatial topological information and mining spatiotemporal correlation. 
This study proposes a Dynamic Fusion Large Language Model (DF-LLM) for traffic flow prediction. The model incorporates three core components: spatiotemporal embedding module, spatiotemporal fusion module, and LLM backbone. The spatiotemporal embedding module enables synergistic representation of multi-scale spatiotemporal features. The spatiotemporal fusion module integrates spatial topology and dynamic dependencies via graph convolution. The LLM backbone adopts a differentiated parameter adaptation strategy to balance training efficiency and traffic data adaptability. Additionally, it introduces a context aggregation attention module to strengthens global dependencies. More importantly, the LLM backbone takes the residual connections to mitigate the gradient vanishing in deep networks. 
Experiments show that DF-LLM has achieved better performance by comparing the metrics on all the four datasets.

\end{abstract}
\section{Introduction}
Within intelligent transportation systems (ITS), traffic flow prediction plays a pivotal role, enabling accurate forecasting of future traffic conditions based on historical data. This capability is essential for advancing traffic management, alleviating congestion, and optimizing resource allocation~\cite{lei2025st,huang2025pte}.

While traditional time series models like ARIMA and Kalman Filtering can capture linear temporal trends, they struggle to model the complex spatiotemporal dependencies in traffic data. Convolutional neural networks (CNNs) and recurrent neural networks (RNNs) have been widely used to capture spatial and temporal dependencies\cite{yuan2018hetero}. Nevertheless, these models are fundamentally constrained by the irregular, non-Euclidean topology of traffic data and its multi-scale periodic patterns. Graph convolutional networks (GCNs)\cite{bai2020adaptive,wu2019graph,yu2018spatio,li2023dynamic} have been introduced to explicitly address spatial structure in traffic networks, but they often suffer from over-smoothing, limiting their ability to capture global patterns. Attention-based models offer more flexibility but come with complex architectures and substantial computational overhead.

LLMs have achieved significant progress across multiple domains and are gradually being utilized in time series research. By leveraging extensive pretraining and massive parameter capacities, LLMs maintain structural stability while enhancing performance\cite{chen2023gatgpt}. However, LLM-based approaches mainly focus on the temporal dimension, largely neglecting the rich spatial topology in traffic networks\cite{caotempo,zhou2023one}. Additionally, the structural and semantic disparities between language and traffic data hinder LLMs from effectively transferring knowledge, potentially compromising prediction performance.

To address these challenges, we propose DF-LLM, a unified spatiotemporal forecasting framework that integrates graph-based spatial modeling with a pretrained Transformer backbone for traffic flow prediction. The proposed framework incorporates a spatiotemporal fusion module with residual connections and a differentiated parameter adaptation strategy, enabling effective domain adaptation while preserving valuable pretrained knowledge. Extensive experiments demonstrate that DF-LLM achieves competitive or superior forecasting performance compared with existing advanced methods, highlighting its potential for intelligent transportation systems.

\section{Proposed Model}
\begin{figure}
\centering
\includegraphics[width=0.95\textwidth, trim=3.5cm 0.4cm 2cm 0.5cm, clip]{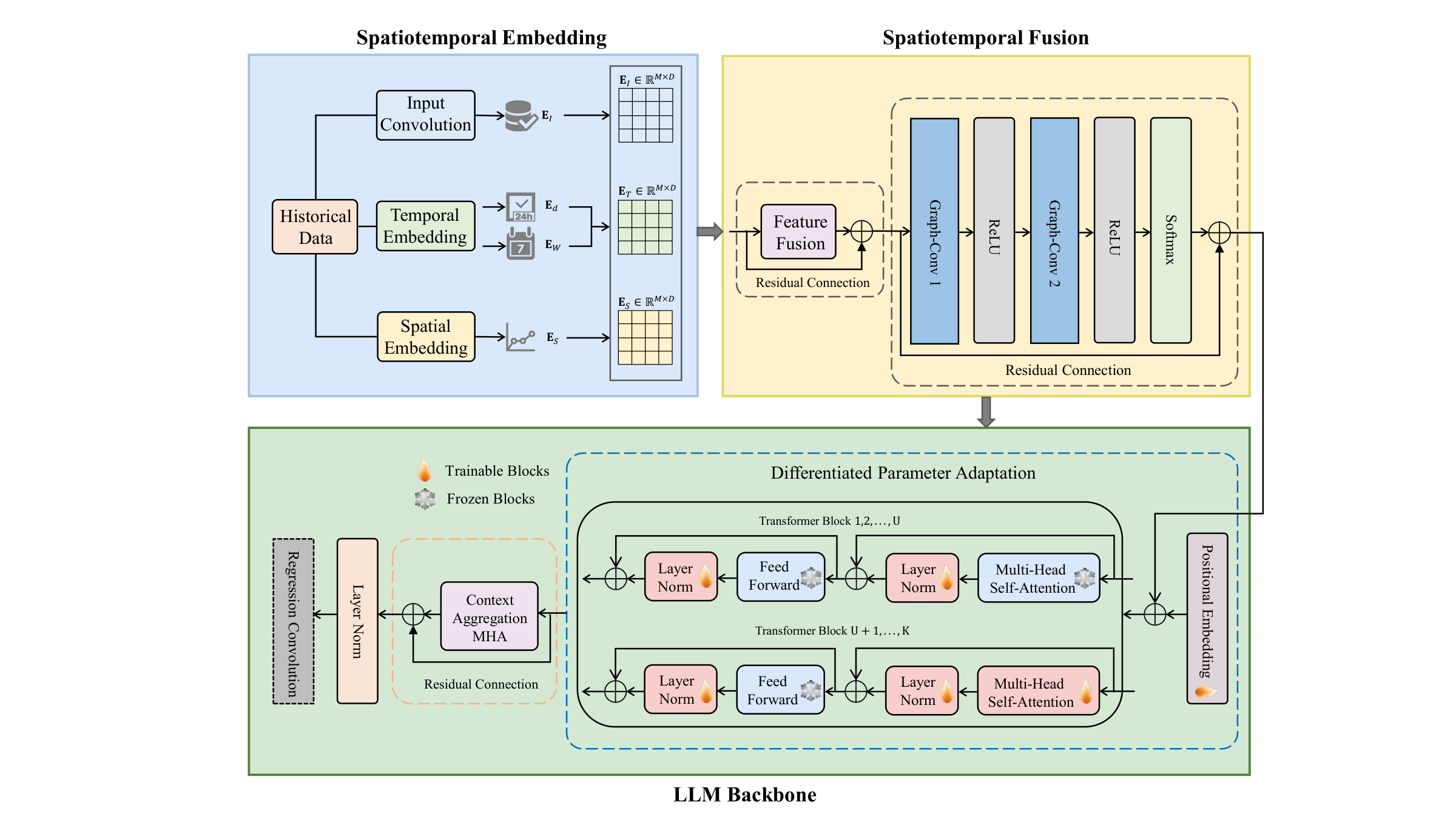}
\caption{The framework of DF-LLM.}
\label{fig1}
\end{figure}

As illustrated in Fig.~\ref{fig1}, the DF-LLM framework is designed to process historical traffic data $\mathbf{X} \in \mathbb{R}^{L \times M \times D}$ as input. The spatiotemporal embedding module generates multi-scale embeddings, including initial, temporal, and spatial feature representations. The spatiotemporal fusion module integrates these embeddings, incorporates residual connections, and leverages a two-layer GCN to model spatial topological relationships, projecting the fused features to dimension $\mathbb{R}^{L \times M \times D'}$. The LLM Backbone integrates positional embedding, Transformer blocks, and a context aggregation multi-head attention (Context Aggregation MHA) module to capture long-range temporal dependencies, adopting a differentiated parameter adaptation strategy that implements partial parameter freezing. Finally, the traffic prediction module outputs the forecasted traffic flow $\widehat{\mathbf{Y}} \in \mathbb{R}^{L' \times M \times D}$ via regression convolution to align with the ground truth $\mathbf{Y} \in \mathbb{R}^{L' \times M \times D}$ for loss calculation.

\subsection{Spatiotemporal Embedding Module}
The spatiotemporal embedding module, which is critical for extracting discriminative representations from historical traffic data, is composed of three parallel branches: initial feature embedding, temporal embedding, and spatial embedding.

\textit{Initial Feature Embedding.} We first employ a pointwise convolution to the input data to extract basic features:
\begin{equation}
\mathbf{E}_I = \text{Conv}_{1 \times 1}(\mathbf{X}_L; \theta_{\text{conv}})
\end{equation}

where $\text{Conv}_{1 \times 1}$ denotes a $1 \times 1$ convolution operation with kernel shape $[1, 1, D, D']$, and $\theta_{\text{conv}}$ represents learnable parameters. This step generates an initial feature embedding $\mathbf{E}_I \in \mathbb{R}^{L \times M \times D'}$.

\textit{Temporal Embedding.} To capture periodic patterns inherent in traffic data, we introduce two learnable parameter matrices: $\mathbf{W}_d$ and $\mathbf{W}_w$. We extract daily and weekly temporal indices $x_d$ and $x_w$ from the input data, then adopt absolute positional encoding to generate $\mathbf{X}_d \in \mathbb{R}^{L \times M \times T_d}$ (daily resolution) and $\mathbf{X}_w \in \mathbb{R}^{L \times M \times T_w}$ (weekly resolution), where $T_d$ and $T_w$ denote the number of time steps per day and per week, respectively. The temporal embedding $\mathbf{E}_T \in \mathbb{R}^{L \times M \times D'}$ is computed as:
\begin{equation}
\mathbf{E}_T = \mathbf{W}_d \cdot \mathbf{X}_d + \mathbf{W}_w \cdot \mathbf{X}_w
\end{equation}

\textit{Spatial Embedding.} To model spatial topological correlations between road segments, we propose an adaptive spatial embedding mechanism:
\begin{equation}
\mathbf{E}_S = \mathcal{E}(W_S \cdot \mathbf{X}_L + b_S)
\end{equation}

Here, $\mathcal{E}$ denotes a non-linear activation function, and the parameters $W_S \in \mathbb{R}^{D \times D'}$ and $b_S \in \mathbb{R}^{D'}$ are learnable. This operation yields a spatial embedding $\mathbf{E}_S \in \mathbb{R}^{L \times M \times D'}$ that captures spatial topological dependencies.

The three embeddings $\mathbf{E}_I, \mathbf{E}_T,$ and $\mathbf{E}_S$ are concatenated and fed into the subsequent spatiotemporal fusion module to achieve multi-scale spatiotemporal feature integration.

\subsection{Spatiotemporal Fusion Module}
This module integrates and refines the embeddings generated by the spatiotemporal embedding module, and captures spatial topological relationships via dynamic graph convolutional operations.

\textit{Feature Fusion.} By concatenating $\mathbf{E}_I, \mathbf{E}_T,$ and $\mathbf{E}_S$ along the feature dimension, we obtain $\mathbf{E}_{\text{concat}} \in \mathbb{R}^{L \times M \times 3D'}$. Subsequently, a $1 \times 1$ convolution layer with residual connections projects this concatenated feature to the target dimension:

\begin{equation}
\mathbf{E}_{\text{fused}} = \text{Conv}_{1 \times 1}(\mathbf{E}_{\text{concat}}; \theta_{\text{fuse}}) + \mathbf{E}_{\text{res}}
\end{equation}

where $\theta_{\text{fuse}}$ denotes learnable parameters of the $1 \times 1$ convolution kernel with a shape of $[1, 1, 3D', D']$, and $\mathbf{E}_{\text{res}} \in \mathbb{R}^{L \times M \times D'}$ is a linear projection of $\mathbf{E}_{\text{concat}}$ to match the output dimension of the $1 \times 1$ convolution layer. Through residual connections, dimensionality reduction from $3D'$ to $D'$ is achieved while preserving essential spatiotemporal features, mitigating the gradient vanishing problem and enhancing the stability of the training process.

\textit{Spatial Topological Modeling with Graph Convolutional Network.} A two-layer GCN with residual connections is employed to $\mathbf{E}_{\text{fused}}$ to capture spatial topological dependencies. Given an adjacency matrix $\mathbf{A} \in \mathbb{R}^{M \times M}$ that characterizes the inherent connectivity of road segments, the GCN operations are defined as:

\begin{equation}
\mathbf{E}_{\text{att}} = \text{Softmax}\left(\text{ReLU}\left(\text{GCN}\left(\text{ReLU}\left(\text{GCN}(\mathbf{E}_{\text{fused}}, \mathbf{A}; \theta_{\text{gcn1}})\right), \mathbf{A}; \theta_{\text{gcn2}}\right)\right)\right)
\end{equation}
\begin{equation}
\mathbf{E}_{\text{sft}} = \mathbf{E}_{\text{att}} + \mathbf{E}_{\text{fused}}
\end{equation}

where $\theta_{\text{gcn1}}$ and $\theta_{\text{gcn2}}$ denote learnable parameters of the two GCN layers, and the GCN operations are applied along the spatial dimension. This operation aggregates feature information from neighboring nodes based on spatial topological correlations, while the residual connection retains direct access to the original fused features. The output $\mathbf{E}_{\text{sft}} \in \mathbb{R}^{L \times M \times D'}$ encapsulates rich multi-scale spatiotemporal information, serving as the input to the LLM Backbone.

\subsection{LLM Backbone}
In this study, GPT-2 is adopted as the LLM backbone for traffic flow prediction through selective parameter adaptation and architectural refinement. Its autoregressive Transformer architecture enables effective temporal dependency modeling while maintaining computational efficiency.

\textit{Differentiated Parameter Adaptation.}
As shown in Fig. \ref{fig1}, the $K$ Transformer blocks of GPT-2 are divided into two subsets with different parameter update strategies. For the first $U$ blocks, the multi-head self-attention (MHA) and feed-forward (FF) layers are frozen to preserve pretrained knowledge, while layer normalization (LN) layers are trainable to adapt to traffic data distributions. For the remaining $(K-U)$ blocks, MHA layers are unfrozen to capture traffic spatiotemporal dependencies, whereas FF layers remain frozen to reduce overfitting. The input and output of GPT-2 are denoted as $\mathbf{E}_{\text{sft}} \in \mathbb{R}^{L \times M \times D'}$ and $\mathbf{E}_{\text{in}} \in \mathbb{R}^{L \times M \times D'}$, respectively.

\textit{Context Aggregation MHA Module.}
To enhance the adapted GPT-2 representations, we introduce a Context Aggregation MHA module to aggregate global spatiotemporal information. By dividing features into $h$ parallel subspaces, this module captures multi-scale correlations and integrates residual connections with layer normalization:

\begin{equation}
\mathbf{E}_{\text{out}}=\text{Norm}(\mathbf{E}_{\text{agg}})+\mathbf{E}_{\text{in}}
\end{equation}

where $\mathbf{E}_{\text{agg}}$ denotes the globally aggregated feature. The resulting representation $\mathbf{E}_{\text{LLM}}$ combines pretrained knowledge with traffic-specific spatiotemporal patterns.

\textit{Traffic Flow Prediction.}
The generated representation $\mathbf{E}_{\text{LLM}}\in\mathbb{R}^{L\times M\times D'}$ is fed into a regression convolution layer to predict future traffic flows:

\begin{equation}
\widehat{\mathbf{Y}}_{L'}=\text{RConv}(\mathbf{E}_{\text{LLM}};\theta_{rc})
\end{equation}

where $\widehat{\mathbf{Y}}\in\mathbb{R}^{L'\times M\times D}$ represents predictions for the next $L'$ time steps and $\theta_{rc}$ denotes the learnable parameters.

\section{Experiments and Results}
\subsection{Experimental Settings}
Experiments are conducted on four public traffic datasets: PEMS04, PEMS08, METR-LA, and PEMS-BAY. Each dataset is split into training, validation, and test sets with a ratio of 6:2:2, and both historical and prediction horizons are set to 12 time steps. DF-LLM is implemented on the BasicTS\cite{shao2024exploring} platform and trained on an NVIDIA GeForce RTX 4090 GPU using the Ranger21 optimizer with a learning rate of 0.0007. The LLM backbone contains six GPT-2 layers, with a batch size of 32 and 500 training epochs; early stopping is applied after 30 epochs without validation improvement. Performance is evaluated using MAE, RMSE, and MAPE, where lower values indicate better accuracy. DF-LLM is compared with representative deep learning methods, including DCRNN\cite{li2018diffusion}, STGCN\cite{yu2018spatio}, GWNet\cite{wu2019graph}, STGODE\cite{fang2021spatial}, STID\cite{shao2022spatial}, AGCRN\cite{bai2020adaptive}, STWave\cite{fang2023spatio}, STAEformer\cite{liu2023spatio}, DGCRN\cite{li2023dynamic}, and STNorm\cite{deng2021st}, as well as LLM-based methods such as ST-LLM\cite{liu2024spatial} and GCNGPT\cite{liu2024spatial}.
\begin{table*}[t]
\centering
\caption{Evaluation results of DF-LLM and baseline methods on four datasets.}
\label{tab2}
\renewcommand{\arraystretch}{1.1}
\setlength{\tabcolsep}{3pt}
\resizebox{\textwidth}{!}{
\begin{tabular}{lcccccccccccc}
\toprule
\multirow{2}{*}{Model} 
& \multicolumn{3}{c}{PEMS04} 
& \multicolumn{3}{c}{PEMS08}
& \multicolumn{3}{c}{METR-LA}
& \multicolumn{3}{c}{PEMS-BAY} \\
\cmidrule(lr){2-4}
\cmidrule(lr){5-7}
\cmidrule(lr){8-10}
\cmidrule(lr){11-13}
& MAE & RMSE & MAPE
& MAE & RMSE & MAPE
& MAE & RMSE & MAPE
& MAE & RMSE & MAPE \\
\midrule
DCRNN     
& 21.72 & 33.95 & 15.74\%
& 17.14 & 27.30 & 10.79\%
& 3.59 & 7.57 & 10.42\%
& 1.97 & 4.60 & 4.72\% \\

STGCN     
& 21.60 & 33.83 & 14.73\%
& 18.25 & 28.55 & 11.73\%
& 3.71 & 7.57 & 10.30\%
& 2.07 & 4.70 & 4.78\% \\

GWNet     
& 20.35 & 32.57 & 14.70\%
& 16.13 & 25.93 & 11.20\%
& 3.55 & 7.41 & 10.00\%
& 1.92 & 4.43 & 4.54\% \\

STGODE    
& 21.06 & 33.31 & 15.24\%
& 17.13 & 26.98 & 11.30\%
& 3.61 & 7.53 & 11.11\%
& 1.98 & 4.48 & 4.66\% \\

STID      
& 19.93 & 32.12 & 13.68\%
& 15.59 & 25.71 & 10.38\%
& 3.54 & 7.50 & 10.84\%
& 1.91 & 4.39 & 4.52\% \\

AGCRN     
& 20.79 & 33.27 & 14.20\%
& 17.53 & 27.90 & 11.28\%
& 3.61 & 7.56 & 10.46\%
& 1.95 & 4.52 & 4.56\% \\

STWave    
& 19.61 & 32.18 & 12.93\%
& 14.84 & 25.48 & 9.81\%
& 3.52 & 7.51 & 10.41\%
& 1.90 & 4.36 & 4.47\% \\

STAEformer 
& 19.40 & 32.23 & 13.08\%
& \textbf{14.76} & 25.57 & \textbf{9.63\%}
& 3.34 & 6.99 & 9.63\%
& 1.91 & 4.35 & 4.50\% \\

DGCRN     
& 20.19 & 32.92 & 14.08\%
& 17.07 & 26.70 & 10.99\%
& 3.44 & 7.23 & 9.90\%
& 1.94 & 4.50 & 4.55\% \\

STNorm    
& 20.46 & 34.20 & 13.68\%
& 16.97 & 27.52 & 10.77\%
& 3.57 & 7.49 & 10.43\%
& 1.94 & 4.50 & 4.64\% \\

GCNGPT  
& 22.94 & 35.26 & 16.55\%
& 18.79 & 29.02 & 12.40\%
& 5.81 & 9.63 & 15.98\%
& 1.78 & 3.87 & 4.06\% \\

ST-LLM   
& 19.74 & 31.31 & 13.81\%
& 15.76 & 24.67 & 10.05\%
& 3.28 & 6.65 & 8.79\%
& 1.67 & 3.72 & \textbf{3.70\%} \\

DF-LLM   
& \textbf{18.96} & \textbf{30.76} & \textbf{12.87\%}
& 15.47 & \textbf{24.60} & 10.11\%
& \textbf{3.12} & \textbf{6.27} & \textbf{8.19\%}
& \textbf{1.63} & \textbf{3.61} & 3.71\% \\
\bottomrule
\end{tabular}
}
\end{table*}

\begin{figure}[h!]
\centering
\includegraphics[width=1.01\textwidth, trim=0cm 0cm 0cm 0cm, clip]{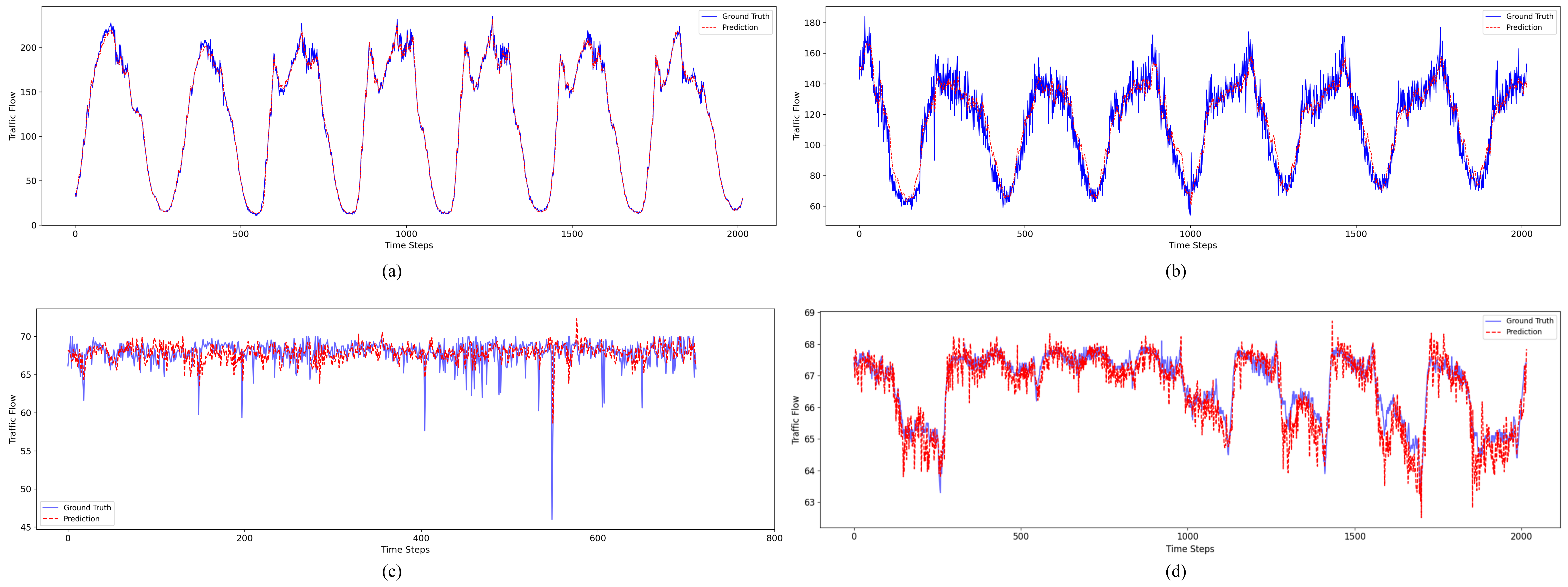}
\setlength{\abovecaptionskip}{-8pt}   
\caption{Comparison of DF-LLM predictions with ground-truth values on four datasets: (a) PEMS04; (b) PEMS08; (c) METR-LA; (d) PEMS-BAY.}
\label{fig8}
\end{figure}
\subsection{Experimental Results}
\textit{Main results.}
The performance comparisons on PEMS04, PEMS08, METR-LA, and PEMS-BAY are summarized in Table \ref{tab2}. Compared with deep learning and LLM-based baselines, DF-LLM achieves competitive forecasting performance across all datasets. On PEMS04, DF-LLM obtains 18.96 (MAE), 30.76 (RMSE), and 12.87\% (MAPE), outperforming most baselines. On PEMS08, it achieves 24.60 (RMSE), demonstrating its ability to capture complex spatiotemporal patterns. For METR-LA, DF-LLM achieves 3.12 (MAE), 6.27 (RMSE), and 8.19\% (MAPE), surpassing ST-LLM and GCNGPT. On PEMS-BAY, it reaches 1.63 (MAE) and 3.61 (RMSE), highlighting the effectiveness of integrating spatial and temporal information.

\textit{Multi-scenario adaptability analysis.}
To further evaluate the robustness of DF-LLM under diverse traffic conditions, visualization results are presented in Fig. ~\ref{fig8}. DF-LLM closely follows the ground truth across different datasets, accurately capturing short-term fluctuations and periodic patterns in PEMS04 and PEMS08, maintaining stable predictions under congested conditions in METR-LA, and modeling complex traffic dynamics in the large-scale PEMS-BAY dataset. Combined with the quantitative results in Table ~\ref{tab2}, these results demonstrate that DF-LLM adapts effectively to diverse traffic patterns and network scales.

\textit{Ablation study.}
Table \ref{tab4} presents ablation results evaluating the contributions of Context Aggregation MHA, GCN, and residual connections (RC). Removing any component leads to performance degradation, while removing residual connections causes the most significant decline. These results verify the effectiveness of the proposed modules and highlight the importance of residual connections in stabilizing model training.

\begin{table}[h!]
\centering
\caption{Ablation study results.}
\label{tab4} 
\makebox[\textwidth][c]{ 
\setlength{\tabcolsep}{0pt} 
\begin{tabular}{l c c c c c c c c c c c c}
\toprule
\multirow{2}{*}{Model} & \multicolumn{3}{c}{PEMS04} & \multicolumn{3}{c}{PEMS08} & \multicolumn{3}{c}{METR-LA} & \multicolumn{3}{c}{PEMS-BAY} \\
\cmidrule(r){2-4} \cmidrule(r){5-7} \cmidrule(r){8-10} \cmidrule(r){11-13}
 & MAE & RMSE & MAPE & MAE & RMSE & MAPE & MAE & RMSE & MAPE & MAE & RMSE & MAPE \\
\midrule
No-MHA   & 19.40 & 31.25 & 13.57\% & 15.53 & 24.73 & 10.20\% & 3.27 & 6.51 &  8.87\% & 1.68 & 3.76 & 3.84\% \\
No-GCN   & 19.09 & 30.90 & 12.94\% & 15.68 & \textbf{24.38} & 10.31\% & 3.23 & 6.45 &  8.74\% & 1.66 & 3.66 & 3.73\% \\
No-RC    & 22.39 & 36.35 & 15.33\% & 19.02 & 30.34 & 12.72\% & 3.36 & 6.70 &  9.02\% & 2.09 & 4.35 & 4.71\% \\
Ours     & \textbf{18.96} & \textbf{30.76} & \textbf{12.87\%} & \textbf{15.47} & 24.60 & \textbf{10.11\%} & \textbf{3.12} & \textbf{6.27} & \textbf{8.19\%} & \textbf{1.63} & \textbf{3.61} & \textbf{3.71\%} \\
\bottomrule
\end{tabular}
}
\end{table}

\section{Conclusion}
This paper proposes DF-LLM, a traffic flow prediction framework that integrates graph convolution with a pretrained LLM backbone for spatiotemporal forecasting. By incorporating spatiotemporal embedding, graph-based spatial modeling, Context Aggregation MHA, and differentiated parameter adaptation, DF-LLM effectively captures complex traffic dynamics. Experiments on four real-world datasets demonstrate competitive forecasting performance compared with existing methods, while ablation studies validate the effectiveness of the proposed components. The model also exhibits stable adaptability across diverse traffic scenarios and network scales, underscoring its potential for intelligent transportation systems. Future work will explore dynamic spatiotemporal modeling and multimodal traffic data to further improve generalization and adaptability.


\bibliographystyle{splncs04}
\bibliography{main}

\end{document}